\documentclass{bmvc2k}

\usepackage{graphicx}
\usepackage{comment}
\usepackage{amsmath,amssymb} 
\usepackage{color}

\newcommand\blfootnote[1]{%
  \begingroup
  \renewcommand\thefootnote{}\footnote{#1}%
  \addtocounter{footnote}{-1}%
  \endgroup
}

\graphicspath{{./figures/}}

\title{Trans2k: Unlocking the Power of Deep Models for Transparent Object Tracking}

\addauthor{Alan Lukežič$^\ast$}{alan.lukezic@fri.uni-lj.si}{1}
\addauthor{Žiga Trojer$^\ast$}{ziga.trojer20@gmail.com}{1}
\addauthor{Jiří Matas}{matas@fel.cvut.cz}{2}
\addauthor{Matej Kristan}{matej.kristan@fri.uni-lj.si}{1}

\addinstitution{
 Faculty of Computer and Information Science \\
 University of Ljubljana \\
 Ljubljana, Slovenia
}
\addinstitution{
 Center for Machine Perception, \\
 Czech Technical University in Prague, \\
 Prague, Czech Republic
}

\runninghead{Lukežič, Trojer, Matas, Kristan}{Trans2k}

\begin{document}

\maketitle

\blfootnote{$^\ast$ The authors contributed equally.}

\begin{abstract}
Visual object tracking has focused predominantly on opaque objects, while transparent object tracking received very little attention. 
Motivated by the uniqueness of transparent objects in that their appearance is directly affected by the background, the first dedicated evaluation dataset has emerged recently.
We contribute to this effort by proposing the first transparent object tracking \textit{training dataset} Trans2k that consists of over 2k sequences with 104,343 images overall, annotated by bounding boxes and segmentation masks. 
Noting that transparent objects can be realistically rendered by modern renderers, we quantify domain-specific attributes and render the dataset containing visual attributes and tracking situations not covered in the existing object training datasets.
We observe a consistent performance boost (up to 16\%) across a diverse set of modern tracking architectures when trained using Trans2k, and show insights not previously possible due to the lack of appropriate training sets.
The dataset and the rendering engine will be publicly released to unlock the power of modern learning-based trackers and foster new designs in transparent object tracking.
\end{abstract}

\section{Introduction}  \label{sec:intro}

Visual object tracking is a fundamental computer vision problem that emerges in a broad range of downstream applications such as human-computer interaction, surveillance, autonomous robotics and video editing, to name a few.
The substantial advances observed in the last decade have been primarily driven by emergence of  challenging evaluation datasets~\cite{otb_pami2015,kristan_vot_tpami2016,got10k,lasot_cvpr19} and  diverse training sets ~\cite{muller_trackingnet,imagenet_ijcv_2015,got10k} that enabled end-to-end learning of modern deep tracking architectures. 
While most benchmarks addressed opaque objects, very little attention has been dedicated to tracking of transparent objects. These are unique in that they are often reflective and their appearance is affected by the background texture, thus reducing the reliability of the deep features trained for opaque objects. 
\begin{figure*}
\centering
\includegraphics[width=\linewidth]{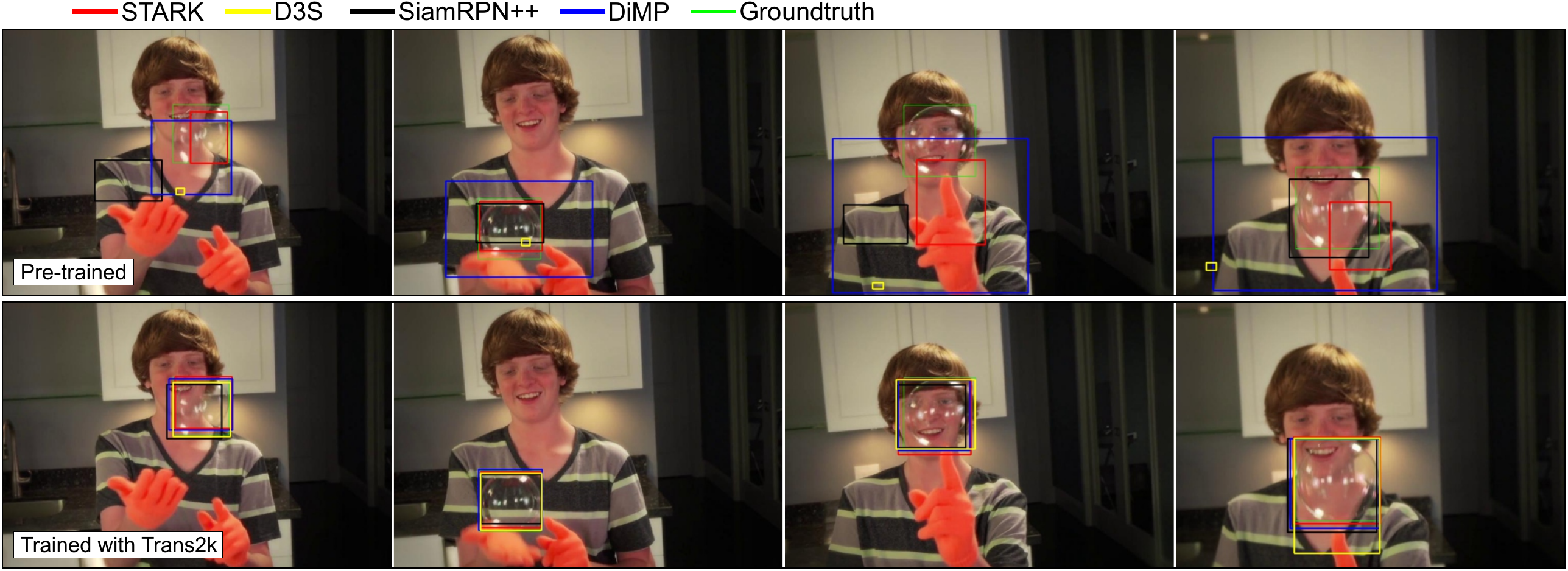}
\caption{
Trackers trained on opaque object training sets fail due to specifics of transparent object appearance dynamics (upper row). After training with the proposed Trans2k, their performance remarkably improves (bottom row).
}
\label{fig:first}
\end{figure*}

Recently, the TOTB benchmark~\cite{totb_iccv21} was proposed to facilitate research in transparent object tracking.
The benchmark results show that classical trackers underperform on transparent objects and that, contrary to opaque object tracking and to many other vision problems, shallow backbones outperform the deep ones. 
However, it is crucial to note that the results were obtained without re-training the state-of-the-art trackers on representative training sets, which opens the question whether these observations are not just a consequence of the domain shift rather than an inherent property of shallow and deep modern learning-based tracking architectures. 
There is thus a pressing need for a high quality transparent object training video dataset to answer this question and to potentially unlock the power of deep learning trackers, as well as to  facilitate in-depth analysis and foster further research. 

Construction of the training dataset presents many challenges.
First, the training set should be large, diverse, and focus on visual attributes and challenging situations specific for transparent objects, which are not already covered in the opaque tracking datasets. 
Second, the targets should be accurately annotated. Presented with these challenges, various sequence selection and annotation protocols have emerged~\cite{got10k,kristan_vot2013,totb_iccv21,kristan_vot2014}. 
In related fields like 6DoF estimation~\cite{Hodan_2020_CVPR,Hodan_2018_ECCV} and scene parsing~\cite{scene_neurips18,sceneparsing_iccv19,kegan_cvpr19}, image rendering has been applied to avoid the aforementioned issues. 
However, the realism of rendered general objects remains limited, reducing the training potential.
We note that transparent objects are unique in that, contrary to their opaque counterparts, non-textured transparent materials may be faithfully rendered by modern renderers~\cite{blenderproc}. 
Thus \textit{highly realistic} sequences with precisely specified visual attributes and pixel-level ground truth free of subjective annotation bias can be generated.

We propose the first transparent object tracking training dataset Trans2k. 
To maximise its utility, a protocol is designed that identifies challenging visual attributes and tracking situations not covered in existing datasets. 
The identified attribute ranges are then used in rendering over 2k sequences with \mbox{104,343} images overall.

A set of trackers representing the major modern deep learning approaches is evaluated on~\cite{totb_iccv21}. We report a consistent performance boost (up to \mbox{$16\%$}) across all architectures when trained with Trans2k.
Contrary to~\cite{totb_iccv21}, we show that deep backbones outperform shallow ones on transparent object tracking, which is consistent with observations in opaque tracking. 
We see transformers as the most promising approach and identify the visual attributes that future architectural designs should address to make significant progress in performance. 

In summary, our contributions are: (i) Trans2k, the first training dataset for transparent object tracking that unlocks the power of deep trainable trackers and allows training bounding box or segmentation trackers, (ii) a complementary analysis on~\cite{totb_iccv21} with new findings indicating future research directions. 
The dataset and the sequence generation engine will be made publicly available. 
The paper reports two surprising observations: first, that transparent object tracking results are comparable to opaque object tracking for state-of-the-art trackers trained with Trans2k, and second, that training with Trans2k leads to substantial performance boost on transparent objects at minimal reduction on opaque objects. 
The dataset, rendering engine and instructions how to use it are available here: \url{https://github.com/trojerz/Trans2k}

\section{Related Work}  \label{sec:related_work}

\textbf{Object tracking.} Deep trackers excel across various benchmarks~\cite{kristan_vot2020,kristan_vot2021,got10k,lasot_cvpr19,otb_pami2015,totb_iccv21} compared to their hand-crafted counterparts.
Initially, pre-trained general backbones were used for feature extraction, primarily by the discriminative correlation filter (DCF) trackers~\cite{danelljan_eccv2016_ccot,DanelljanCVPR2017,danelljan_eccv2018_updt,danelljan_iccvw2015,liu_icme2021}, which learned a discriminative localization models online during tracking.
Later, backbone end-to-end training techniques that maximize DCF localization were proposed~\cite{Valmadre_2017_CVPR}. Most recently, the DCF optimization has been introduced as part of the deep network. Milestone representatives were proposed in~\cite{atom_cvpr19,danelljan_dimp_iccv19}, which also 
proposed a post-processing network for bounding box refinement that accounted for target aspect changes.
In parallel, siamese trackers have been explored and grown into a major tracker design branch. The seminal work~\cite{siamfc_eccvw2016} trained AlexNet-based network~\cite{alexnet_nips12} such that localization accuracy is maximized simply by correlation between a template and search region in feature space. 
These trackers afford fast processing since no training is required during tracking. Siamese trackers were extended by anchor-based region proposal networks~\cite{siamrpn_cvpr2018,siamrpn_cvpr2019} and recently an anchor-free extension has been proposed~\cite{siamban_cvpr20} with improved localization performance. 
Drawing on advances in object detection~\cite{detr_eccv2020}, transformer-based trackers have recently emerged ~\cite{stark_iccv21,transt_cvpr2021,transf_cvpr2021}.
These are the current state-of-the-art, and computationally efficient with remarkable real-time performance~\cite{kristan_vot2021}. 

\textbf{Benchmarks.} The developments in visual object tracking have been facilitated by introduction of benchmarks. The first widely-used benchmark~\cite{otb_pami2015,otb_cvpr2010} proposed a dataset and evaluation protocol that allowed standardised comparsion of trackers.
Later, the VOT initiative explored dataset construction as well as performance evaluation protocols for efficient in-depth analysis~\cite{kristan_vot_tpami2016,kristan_vot2013,kristan_vot2014}. Further improvements were made in the subsequent yearly challenges, e.g.,~\cite{kristan_vot2020,kristan_vot2021}. 
With advent of deep learning, tracking training sets have emerged.
\cite{muller_trackingnet} constructed a huge training set from public video repository and applied a semi-automatic annotation. 
Recently,~\cite{got10k} presented ten thousand annotated video sequences, divided into a large training and a smaller evaluation set. 
Concurrently, a long-term tracking benchmark~\cite{lasot_cvpr19} with fifteen pre-defined categories, containing training and test set was proposed.  
All these benchmarks focus on opaque objects, while recently as transparent object tracking evaluation dataset ~\cite{totb_iccv21} has been proposed. However, training datasets for transparent object tracking have not been proposed.

\textbf{Use of synthesis.} 
Rendering has been previously considered in computer vision to avoid costly manual dataset acquisition. In~\cite{video_games_cvpr2018,playing_iccv2017}, synthetic data was generated by a video game engine, which provided an unlimited amount of annotated training data for various computer vision tasks.
A rendered dataset of urban scenes, Synthia~\cite{synthia_cvpr2016}, was shown to substantially improve the trained deep models for semantic segmentation. 
A similar dataset~\cite{synscapes_2018} was proposed for training and evaluation of scene parsing networks.
A fine-grained vegetation and terrain dataset~\cite{metzger_icpr2021} was recently proposed for training drivable surfaces and natural obstacles detection networks in outdoor scenes.
\cite{synthetic_eccv2018} showed that foreground and background should be treated differently when training segmentation on synthetic images.
The benefits of using mixed real and synthetic 6DoF training data has been recently shown in~\cite{hodan_cvpr2020}.
The major 6DoF object detection challenge~\cite{hodan_bop2020} thus provides a combination of real and synthetic images for training as well as evaluation.  
Synthesis has been used in the UAV123 tracking benchmark~\cite{uav_benchmark_eccv2016} in which eight of the sequences are rendered by a game engine. 
A rendering approach was used in~\cite{cehovin_iccv2017}  to parameterize camera motion for fine-grained tracker performance analysis.
However, using synthetic data for training in visual tracking remains unexplored.
 
\textbf{Transparent objects.} Highlighting the difference from opaque counterparts, transparent objects have been explored in computer vision in various tasks. 
Recognition of transparent objects was studied in~\cite{fritz_nips2009,maeno_cvpr2013}, while 3D shape estimation and reconstruction of transparent objects on RGB-D images was proposed in~\cite{klank_icra2011,sajjan_icra2020}. 
Segmentation of transparent objects has been studied in~\cite{xu_iccv2015,kalra_cvpr2020}, while a benchmark was proposed in~\cite{xie_eccv2020}. 
All these works consider single-image tasks and little attention has been dedicated to videos. 
In fact, a transparent object tracking benchmark~\cite{totb_iccv21} has been proposed only recently and reported a performance gap between transparent and opaque object tracking. 
However, due to the lack of a dedicated training dataset, the gap source remains unclear.

\section{Trans2k dataset}  \label{sec:dataset}

Transparent objects, which are often reflective and glass-like, can be rendered with a high level of realism by the modern photo-realistic rendering engines~\cite{blenderproc}.
In our approach, we first identify and parameterize the sequence attributes specific to transparent objects (Section~\ref{sec:attributes}). 
A BlenderProc-based sequence generator is implemented that enables parameterized sequence rendering. 
Attribute levels useful for learning are identified empirically and the final training dataset is generated (Section~\ref{sec:range_identification}).

\subsection{Parametrization of sequence attributes}  \label{sec:attributes}

The dataset should reflect the diversity of visual attributes typical for transparent object tracking scenes for efficient learning. After carefully examining various videos of transparent and opaque objects, the following attributes were identified (Figure~\ref{fig:parameter_examples}).  

{\noindent \bf Scene background.} Since background affects the transparent object appearance, a high background diversity is required in training. We ensure this by randomly sampling videos from GoT10k~\cite{got10k} training set and use them as backgrounds over which the transparent object is rendered.

{\noindent \bf Object types.} 3D models of 25 object types from open source online repositories are selected with several instances of the same type. 
The set was chosen such to cover a range of nontrivial as well as smooth shapes, with some objects rendered with empty and some with full volume. This amounts to 148 object instances. 

{\noindent \bf Target motion.} To increase the object-background appearance diversity, the objects are moving in the videos. The motion trajectory is generated by a cubic Hermite spline spanned by four uniformly sampled points. The motion dynamics is not critical in training, since deep models are typically trained on pairs of image patches cropped at target position. Thus a constant velocity is applied.

{\noindent \bf Distractors.} In realistic environments, the target may be surrounded by other visually similar transparent objects (e.g., glasses on a table), which act as distractors. We thus render an additional transparent object following the target object. 
The distractor object is from a different type to keep the appearance-based localization learning task feasible.

{\noindent \bf Transparency.} The level of transparency crucially affects the target appearance. We thus identify four levels ranging from clearly visible to nearly invisible.

{\noindent \bf Motion blur.} Fast motions, depending on the aperture speed, result in various levels of blurring. We identify four levels of blur intensity, ranging from no blurring to extreme blurriness.

{\noindent \bf Partial occlusion.} Objects are commonly occluded by other objects in practical situations (e.g., handling of the target). We thus simulate partial occlusions by rendering coloured stripe pattern moving across the video frame. The stripe width is fixed, while the occlusion intensity is simulated by the number of stripes (0, 7, 11, 20) per image, i.e., from zero to severe occlusion.

{\noindent \bf Rotation.} To present realistic object appearance change, the object rotates in 3D in addition to position change. The rotation dynamics is specified by the angular velocity along each axis, which is kept constant throughout the sequence. We identify four rotation speed levels, (0, 1.3, 5.4, 10.6) degrees per frame, thus ranging from no rotation to fast rotation.  

\subsection{Attribute selection and dataset generation}  \label{sec:range_identification}

\begin{figure*}[t]
\centering
\includegraphics[width=\linewidth]{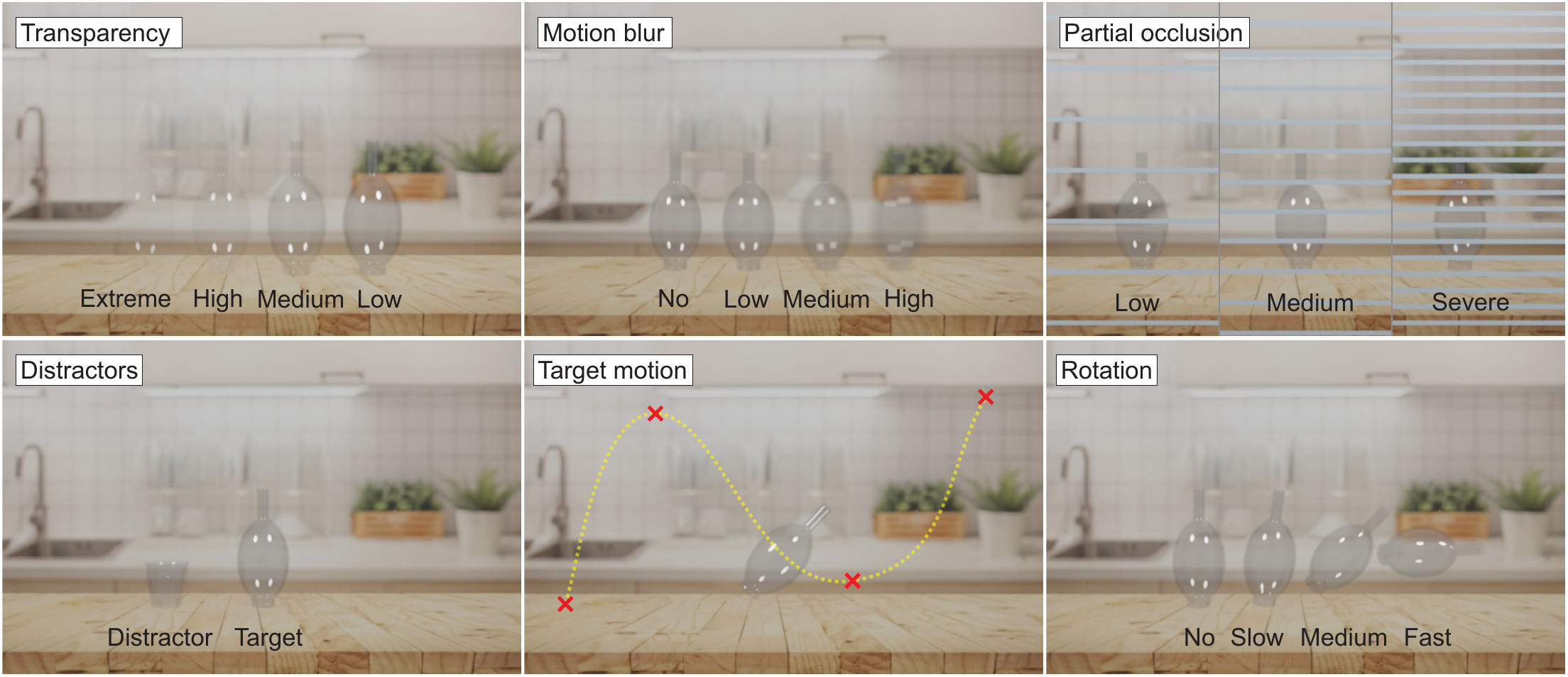}
\caption{Trans2k attribute levels for "Transparency", "Motion blur", "Partial occlusion", "Distractor" (binary), "Target motion" (four control points) and "Rotation". 
}
\label{fig:parameter_examples}
\end{figure*}

To maximize the dataset application utility, the sequences should be complementary to existing datasets from tracking perspective and should focus on attributes that the learning-based trackers cannot already learn from opaque object tracking training sets. An empirical study was designed to determine which intensity levels of the attributes (i) transparency, (ii) partial occlusion, (iii) rotation and (iv) motion blur should be considered in the final dataset. The intensity levels are visualized in Figure~\ref{fig:parameter_examples}.
 
Seven state-of-the-art deep learning trackers pretrained on opaque object tracking datasets (see Section~\ref{sec:experiments} for details) that cover the major current trends in tracking were selected. The difficulty level of individual attribute is quantified as the overall performance of these trackers on test sequences rendered with that attribute. Specifically, for each attribute level, five  sequences were sampled from Got10K~\cite{got10k} training set and used as backgrounds in the rendered sequences. The same background sequences with same object, traveling over the same trajectory are used with all attributes to ensure consistent evaluation. This resulted in 80 test sequences (4 attributes $\times$ 4 levels $\times$ 5 variations). 

\begin{figure*}[t]
\centering
\includegraphics[width=\linewidth]{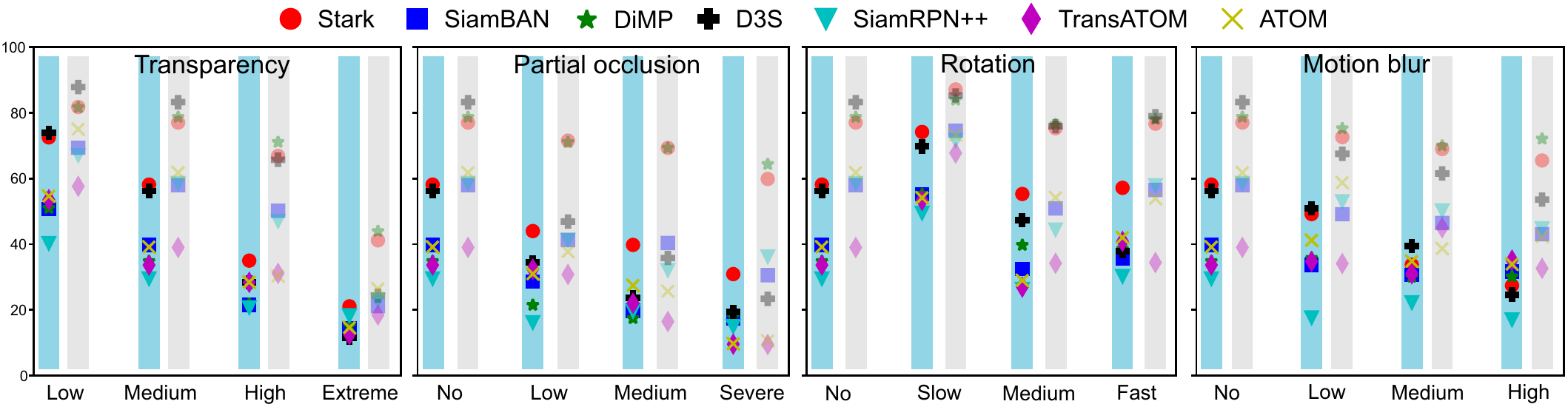}
\caption{Average IoU of trackers reflect the difficulty level of individual attribute intensity. The blue shaded columns show performance of trackers pre-trained on opaque datasets, while the gray shaded column shows performance after training with Trans2k.}
\label{fig:parameters_experiment}
\end{figure*}

The results are shown in Figure~\ref{fig:parameters_experiment}. We observe that most of the attribute levels result in performance reduction and are thus kept as relevant in our final dataset, except from two at which the trackers score quite high. 
The lowest transparency level and zero rotation appear to be well addressed by the opaque object training sets, thus we decide not to include them in our dataset for better use of its capacity. 
The following parameters are thus applied when rendering Trans2k. The GoT10k training set sequences are sampled at random and at most once. All object types are sampled with equal probability. The transparency levels (excluding the lowest level) are sampled with equal probability. Blur presence in a sequence is sampled with 0.15 probability, with blur levels sampled uniformly. Occlusion presence is sampled with 0.2 probability, while occlusion levels are sampled uniformly. Rotation level is uniformly sampled. The resulting training dataset Trans2k thus contains 2,039 challenging sequences and 104,343 frames in total.

Since the sequences are rendered, the ground truth can be exactly computed. 
We provide the ground truth in two standard forms, the widely accepted target enclosing axis-aligned bounding-box and the segmentation mask to cater to the emerging segmentation trackers~\cite{kristan_vot2020}. 
The ground truths for distractors are generated as well.  
Trans2k is thus the first dataset with per-frame distractor annotation to facilitate development of future learning-based methods that could exploit this.   

\section{Experiments}  \label{sec:experiments}
 
\subsection{Selected trackers and training setup}  \label{sec:trackers}

We selected state-of-the-art learning-based trackers that cover the major trends in modern architecture designs for validating Trans2k:
(i) two siamese trackers SiamRPN++~\cite{siamrpn_cvpr2019} and SiamBAN~\cite{siamban_cvpr20}, 
(ii) two deep correlation filter trackers ATOM~\cite{atom_cvpr19} and DiMP~\cite{danelljan_dimp_iccv19}, 
(iii) the recent state-of-the-art transparent object tracker TransATOM~\cite{totb_iccv21}, and 
(iv) a transfomer-based tracker STARK~\cite{stark_iccv21}. 
These trackers localize the target by a bounding box. To account for the recent trend in localization by per-pixel segmentation~\cite{kristan_vot2020}, we include 
(v) the recent state-of-the-art segmentation-based tracker D3S~\cite{Lukezic_CVPR_2020}.

During training, the trackers were initialized by the pre-trained weights provided by their authors, while all the training details were the same as in the original implementations. 
The trackers were trained for 50 epochs with 10000 training samples per epoch. Since Trans2k was designed as a complementary dataset covering situations not present in existing datasets, the training considers samples from Trans2k as well as opaque object sequences. In particular, we merged the opaque training datasets GOT10k~\cite{got10k}, LaSoT~\cite{lasot_cvpr19} and TrackingNet~\cite{muller_trackingnet} into a single dataset, abbreviated as opaque object training dataset (OTD). A training batch is then constructed by sampling from Trans2k and OTD with 5:3 ratio.

\subsection{Validation of Trans2k} \label{sec:baseline_comparison}

We first validated the contribution of Trans2k by measuring performance of trackers on the recent transparent object tracking benchmark TOTB~\cite{totb_iccv21}. 
Following the regime described in Section~\ref{sec:trackers} the selection of seven state-of-the-art trackers was trained using Trans2k. 
Their performance was then compared to their original performance, i.e., when trained only with opaque object tracking sequences. Thus any change in performance is contributed only by the training dataset.
The trackers were evaluated by the standard one-pass evaluation protocol (OPE) that quantifies the performance by AUC and center error measures on success and precision plots. For more information on the protocol, please refer to~\cite{otb_pami2015,totb_iccv21}. 
\begin{figure*}[t]
\centering
\includegraphics[width=\linewidth]{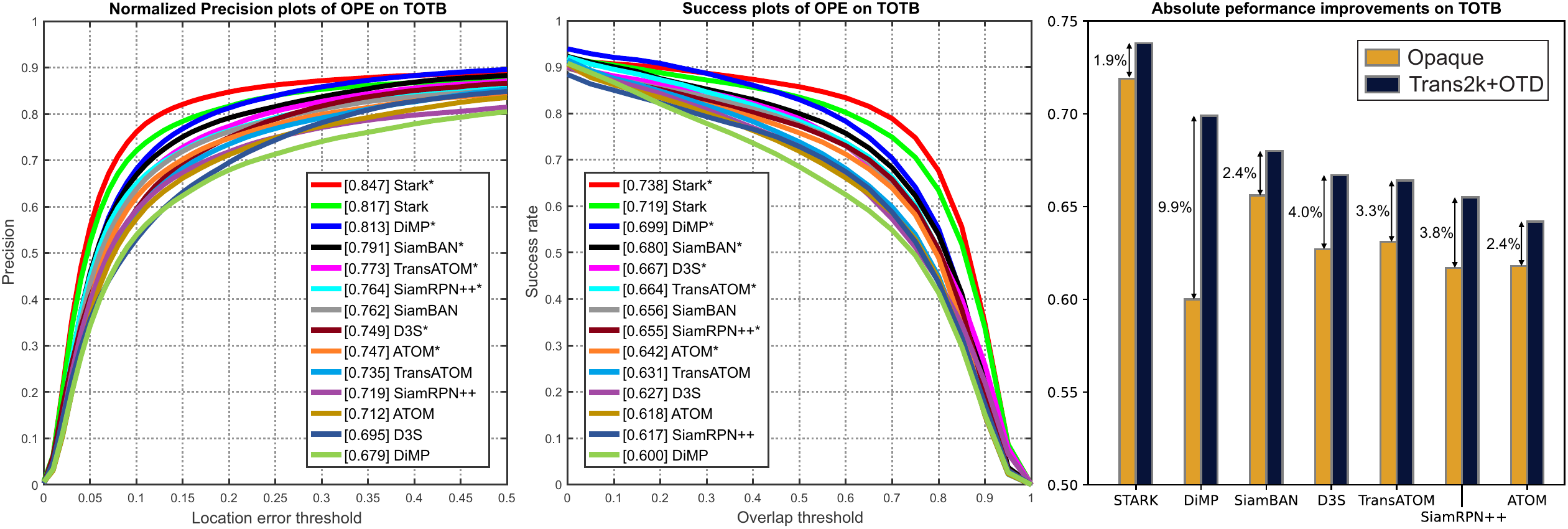}
\caption{Trackers evaluated on TOTB dataset shown in precision and success plots. Trackers trained with Trans2k are denoted by a star (*). The right graph shows absolute improvements in tracking performance measured by the AUC measure after training with the proposed Trans2k.}
\label{fig:baseline_comparison}
\end{figure*}

The results are shown in Figure~\ref{fig:baseline_comparison}. The performance of all trackers substantially improved when trained using Trans2k. The performance gains are at a level usually expected for a clear methodological improvement. 
Recently, TransATOM~\cite{totb_iccv21}, a transparent object tracking extension of ATOM~\cite{atom_cvpr19}, was proposed, which outperformed ATOM by 2.1\%. 
Without any methodological modification and only training with Trans2k, ATOM \textit{outperforms} this extension by 1.7\%. Nevertheless, TransATOM gains 3.3\% when trained with Trans2k.
The largest performance boost is achieved by DiMP, which improves by over 16\% and scores as the second-best among all the tested trackers.
Consistent with the observation on opaque object tracking benchmarks, the transformer-based tracker STARK achieves the best performance. Note that even without training with Trans2k, STARK surpasses all trackers, but when trained with Trans2k, an additional healthy 2.5\% performance boost is observed.
Since Trans2k provides segmentation ground truths in addition to bounding boxes, it boosts the segmentation-based tracker D3S~\cite{Lukezic_CVPR_2020} as well. The version trained with Trans2k gains a remarkable 6\% in performance.

\subsection{Re-evaluating the significance of backbone depth}  \label{sec:backbone_depth_experiment}

The recent benchmark~\cite{totb_iccv21} reported a remarkable case that, specific to transparent object tracking, shallow backbones outperform deep ones, which conflicts common observations in opaque object tracking. 
Since~\cite{totb_iccv21} could only analyze the performance with using opaque tracking training sets, we re-evaluate this claim but in the context of using a transparent object tracking training set. 
We select three deep discriminative correlation filters DiMP~\cite{danelljan_dimp_iccv19}, ATOM~\cite{atom_cvpr19} and TransATOM~\cite{totb_iccv21} and study their performance with a shallow (ResNet18) and a deep (ResNet50) backbone when trained with and without transparent objects.

Results in Table~\ref{tab:backbone_depth} reveal that ATOM and TransATOM with shallow backbones indeed outperform their deep backbone counterparts. 
In contrast, DiMP with deep backbone substantially outperforms the shallow backbone counterpart. 
This apparent discrepancy comes from the different designs of ATOM and DiMP. While ATOM uses a pre-trained backbone and allows training of only post-processing steps, DiMP trains the backbone as well. 
The ATOM's apparent preference of shallow backbones comes from the fact that shallow backbones generalize better to transparent objects when trained only with opaque training examples. 
To verify this, we replace ATOM's and TransATOM's backbone by those trained by DiMP. Both trackers improve their performance with deep backbones trained on transparent objects compared to shallow ones. 
Interestingly, their deep backbone variants reach performance near DiMP's and their \textit{performance difference becomes negligible} -- thus properly trained vanilla ATOM should be preferred to its more complex extension TransATOM.
The experiments thus reveal, that deep backbones in fact lead to substantial improvements over shallow counterparts, if trained on the transparent object dataset. 
\setlength{\tabcolsep}{4pt}
\begin{table}[h]
\begin{center}
\caption{Tracking performance (AUC) of three trackers using different backbones. \textit{Opaque} indicates training with only OTD, \textit{+ Trans2k} to using the transparent dataset as well.
Pretrained ResNet18 and ResNet50 backbones are denoted by R18 and R50, respectively, while their versions trained by DiMP are denoted by D18 and D50.}
\label{tab:backbone_depth}
\begin{tabular}{l|cc|cccc|cccc}
\hline
 \multicolumn{1}{c}{ } & \multicolumn{2}{c}{DiMP} & \multicolumn{4}{c}{ATOM} & \multicolumn{4}{c}{TransATOM} \\
  & D18 & D50 & R18 & R50 & D18 & D50 & R18 & R50 & D18 & D50 \\
\hline
Opaque & 0.552 & 0.600 & 0.618 & 0.608 & 0.551 & 0.588 & 0.631 & 0.608 & 0.582 & 0.603 \\
+ Trans2k & 0.613 & 0.699 & 0.642 & 0.648 & 0.629 & 0.695 & 0.664 & 0.664 & 0.647 & 0.697 \\
\hline
\end{tabular}
\end{center}
\end{table}

\subsection{How does Trans2k affect opaque object tracking?}\label{sec:opaque_tracking}

To quantify how much the trackers trained with Trans2k lose in generalization to opaque objects, we evaluate the trackers on the GOT10k~\cite{got10k} validation dataset. 
Table~\ref{tab:gotval_results} shows results for trackers trained only with OTD and with added Trans2k (as described in Section~\ref{sec:trackers}). 
The tracking performance on opaque objects slightly drops, but still remains high. 
This result suggests that, while substantial boosts are observed in transparent object tracking (Figure~\ref{fig:baseline_comparison}) with the use of Trans2k, the generalization to opaque objects is not lost. 
\setlength{\tabcolsep}{3pt}
\begin{table}[h]
\begin{center}
\caption{Tracking performance (AUC) on the opaque tracking dataset GoT-10k val. \textit{Opaque} -- training with only OTD, \textit{+ Trans2k} -- using the transparent dataset as well.
}
\label{tab:gotval_results}
\begin{tabular}{lccccccc}
\hline
  & STARK & DiMP & SiamBAN & D3S & TransATOM & SiamRPN & ATOM \\
\hline
Opaque & 0.777 & 0.706 & 0.679 & 0.676 & 0.662 & 0.656 & 0.650 \\
+ {Trans2k}  & 0.752 & 0.696 & 0.676 & 0.663 & 0.650 & 0.656 & 0.650 \\
\hline
\end{tabular}
\end{center}
\end{table}

\subsection{The role of using opaque objects in training}\label{sec:role_of_opaque}

To further study the impact of the training set content from perspective of the presence of opaque and transparent objects, the training sets were varied. 
We selected two well-known state-of-the-art trackers that performed well in our previous experiments, yet could be trained sufficiently fast. 
The deep discriminative correlation filter DiMP~\cite{danelljan_dimp_iccv19} and the siamese tracker SiamBAN~\cite{siamban_cvpr20} were selected. 
The original versions trained by the authors were evaluated on TOTB~\cite{totb_iccv21} along with the versions re-trained using the following variations of the training set: 
(i) only Trans2k without OTD, 
(ii) only OTD, 
(iii) Trans2k+OTD. 
In experiments (i), (ii) and (iii) the tracker networks are initialized by their pre-trained models provided by the authors. 
Thus an additonal experiment (iv) is performed where the trackers were trained from scratch using the dataset from (iii). 
The trained trackers were evaluated on TOTB~\cite{totb_iccv21}.

Results in Table~\ref{tab:training_ablation} show that using only Trans2k reduces the tracking performance compared to training on opaque objects training datasets. 
A closer look revealed that the trackers trained only with Trans2k tend to focus on transparent objects in general rather than localizing the target, which was reflected in tracker often jumping to nearby transparent objects when multiple such objects were close to the target.
Training with OTD when initialized with original tracker parameters does not bring improvements in general (DiMP performance drops slightly, while that of SiamBAN increases a bit). 
However, when using transparent \textit{as well as} opaque objects in the training set, the performance improves substantially. 
When training from scratch, the performance of both trackers drops compared to the version initialized with pre-trained networks. 
This suggests pre-training is beneficial for both trackers, but particularly for SiamBAN as it requires learning more parameters than DiMP.
\setlength{\tabcolsep}{4pt}
\begin{table}[h]
\begin{center}
\caption{
Comparison of different training setups for SiamBAN and DiMP. 
Performance of pre-trained trackers is indicated by \textit{orig.}, while \textit{scr.} indicates training from scratch.
}
\label{tab:training_ablation}
\begin{tabular}{l|c|cccc}
\hline
  & orig. & Trans2k & OTD & OTD+Trans2k & scr. \\
\hline
DiMP & 0.600 & 0.554 & 0.584 & 0.699 & 0.667 \\
SiamBAN & 0.656 & 0.650 & 0.658 & 0.680 & 0.649 \\
\hline
\end{tabular}
\end{center}
\end{table}

\section{Conclusion}  \label{sec:conclusion}

The first transparent object tracking training dataset Trans2k is proposed. The fact that transparent objects can be sufficiently realistically rendered by modern renderers is exploited. 
Using a specialized protocol, we identified visual attributes not covered well in existing datasets and rendered a dataset with over 2k training sequences containing transparent objects. 

Trans2k was validated on the recent transparent object tracking benchmark TOTB~\cite{otb_pami2015}. Training with Trans2k improves performance at levels usually observed in fundamental methodological advancements in tracking algorithms. 
This behavior is observed over a wide range of tracking methodologies. 
Analysis shows that significant performance gains in transparent object tracking come at a minor performance loss in opaque object tracking, which indicates to excellent generalization of modern trackers.
In contrast to the findings in~\cite{totb_iccv21}, experiments show that trackers benefit from training deeper backbones on transparent objects. 
Additional experiments showed the benefits of using transparent as well as opaque objects in the training dataset. 
Overall, the best performance was observed with transformers.

While the field of transparent object tracking has recently obtained an excellent test set~\cite{totb_iccv21}, the main ingredient crucial for advancements, i.e., a curated training set was missing. 
Trans2k fills this void and will enable future development of new learnable modules specifically addressing the challenges in transparent object tracking, thus fully unlocking the  power of modern deep learning trackers on this scientifically interesting domain. 
Our sequence generator engine will be released along with Trans2k. 
We envision that the engine will allow inovative learning modes in which the sequences with specific challenges can be generated on demand to specialize the trackers to niche tasks or to improve their overall performance. In addition, the rendering 
engine could be used to generate training data for 6-DoF video pose estimation, thus benefiting research beyond 2D transparent object tracking.

\hfill \break

\noindent {\bf Acknowledgements}: 
This work was supported by Slovenian research agency program 
P2-0214 
and projects 
Z2-4459, 
J2-2506 
and 
J2-9433. 
J. Matas was supported by the Czech Science Foundation grant GA18-05360S. 

\bibliography{egbib}

\clearpage
\appendix

\section{Per-attribute analysis}  \label{sec:per_attribute}

\begin{figure*}[h]
\centering
\includegraphics[width=\linewidth]{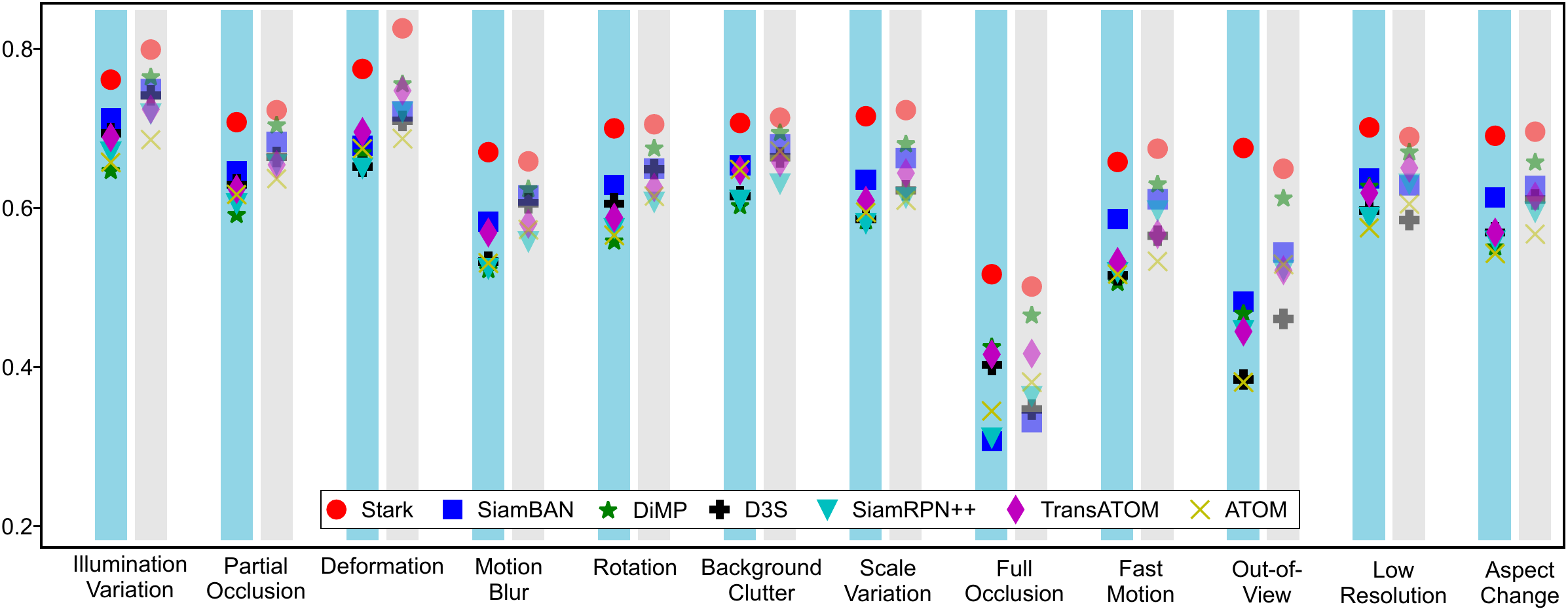}
\caption{Per-attribute performance on TOTB~\cite{totb_iccv21}. Blue ribbons indicate trackers trained with opaque training set, while gray indicate performance after training with Trans2k.}
\label{fig:per-attribute}
\end{figure*}

Figure~\ref{fig:per-attribute} shows per-attribute tracking performance on TOTB dataset~\cite{totb_iccv21} with and without training on Trans2k.
Most of the trackers improve at all attributes when trained with Trans2k, which leads to a consistent average tracker per-attribute boost. Interestingly, after training with Trans2k, the per-attribute difference between trackers decreases, thus reducing the influence of the methodological differences. The per-attribute tracker ranks change, but STARK~\cite{stark_iccv21}
remains the top tracker in all attributes.
The most difficult attributes remain full occlusion and out-of-view, which is expected since the short-term trackers are not specifically designed for target re-detection, which is a long-term property. 
In addition, the fast motion and motion blur appear to be the next most difficult attributes. These are the attributes future research should address in transparent object tracking. 
The least challenging attributes are illumination variation and deformation, which are apparently well addressed by the powerful deep backbone features. Partial occlusion also seems to be addressed well by the existing trackers.

\section{The role of training set size}  \label{sec:role_of_dsetsize}

To study the required number of transparent object training sequences in the training regime, we re-trained DiMP~\cite{danelljan_dimp_iccv19} and SiamBAN~\cite{siamban_cvpr20} with varying the training dataset size. 
In particular, the number of transparent object sequences in the training set (originally 2,039 sequences) was reduced to 100, 500 and 1000. 

The results in Table~\ref{tab:training_ablation} show that using only 100 sequences already significantly improves the results for DiMP, while SiamBAN requires at least 500 sequences to achieve a noticeable improvement over its pre-trained version on opaque datasets. 
We speculate that the reason is that SiamBAN relies more on the dataset diversity since it needs to learn features capable of object localization only by template correlation. 
On the other hand, the DiMP backbone only needs to learn features that sufficiently well reflect the transparent object presence, while the discriminative template is learned online. 
However, DiMP further improves with the additional sequences in the full Trans2k dataset.

\setlength{\tabcolsep}{4pt}
\begin{table}
\begin{center}
\caption{
Comparison of different sizes of the training dataset Trans2k for SiamBAN and DiMP. 
The numbers 100, 500, 1000 and 2039 denote number of sequences containing transparent objects, which were used for training.
Performance of trackers pre-trained on opaque objects only is indicated by \textit{orig.}
}
\label{tab:training_ablation}
\begin{tabular}{l|c|cccc}
\hline
  & orig. & 100 & 500 & 1000 & 2039 \\
\hline
DiMP & 0.600 & 0.649 & 0.679 & 0.687 & 0.699 \\
SiamBAN & 0.656 & 0.655 & 0.667 & 0.670 & 0.680 \\
\hline
\end{tabular}
\end{center}
\end{table}

\section{Qualitative examples}  \label{sec:qualitative}

\begin{figure*}[h]
\centering
\includegraphics[width=\linewidth]{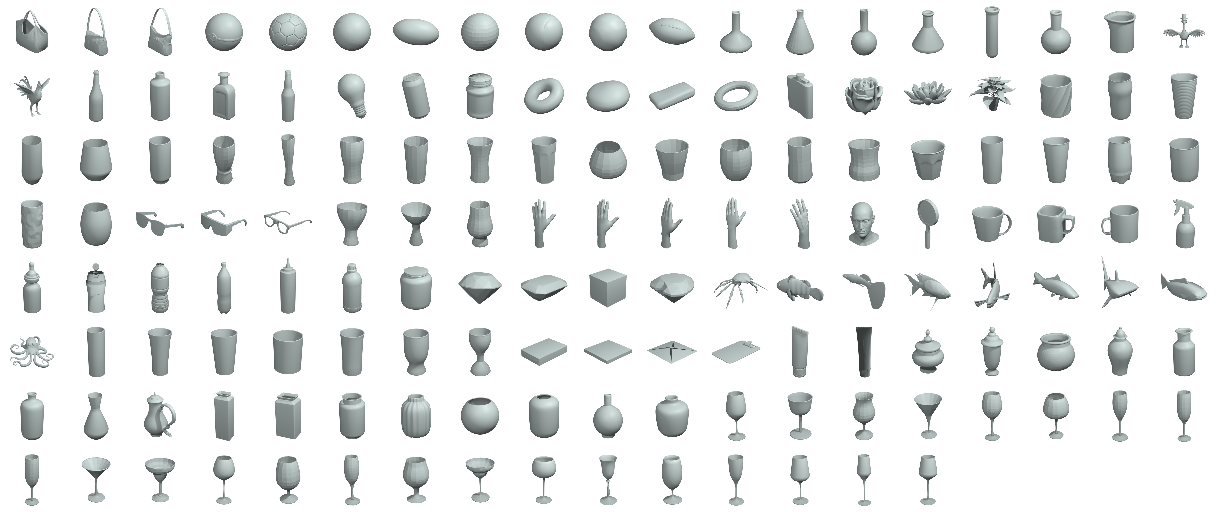}
\caption{A diverse set of object instances used in Trans2k sequences.}
\label{fig:object_types}
\end{figure*}

In this section we present some qualitative examples of the proposed dataset.
Figure~\ref{fig:object_types} shows objects which was used for rendering the Trans2k. 
We defined 25 object types and each object type has several instances, which results in total 148 object instances. 

\begin{figure*}
\centering
\includegraphics[width=\linewidth]{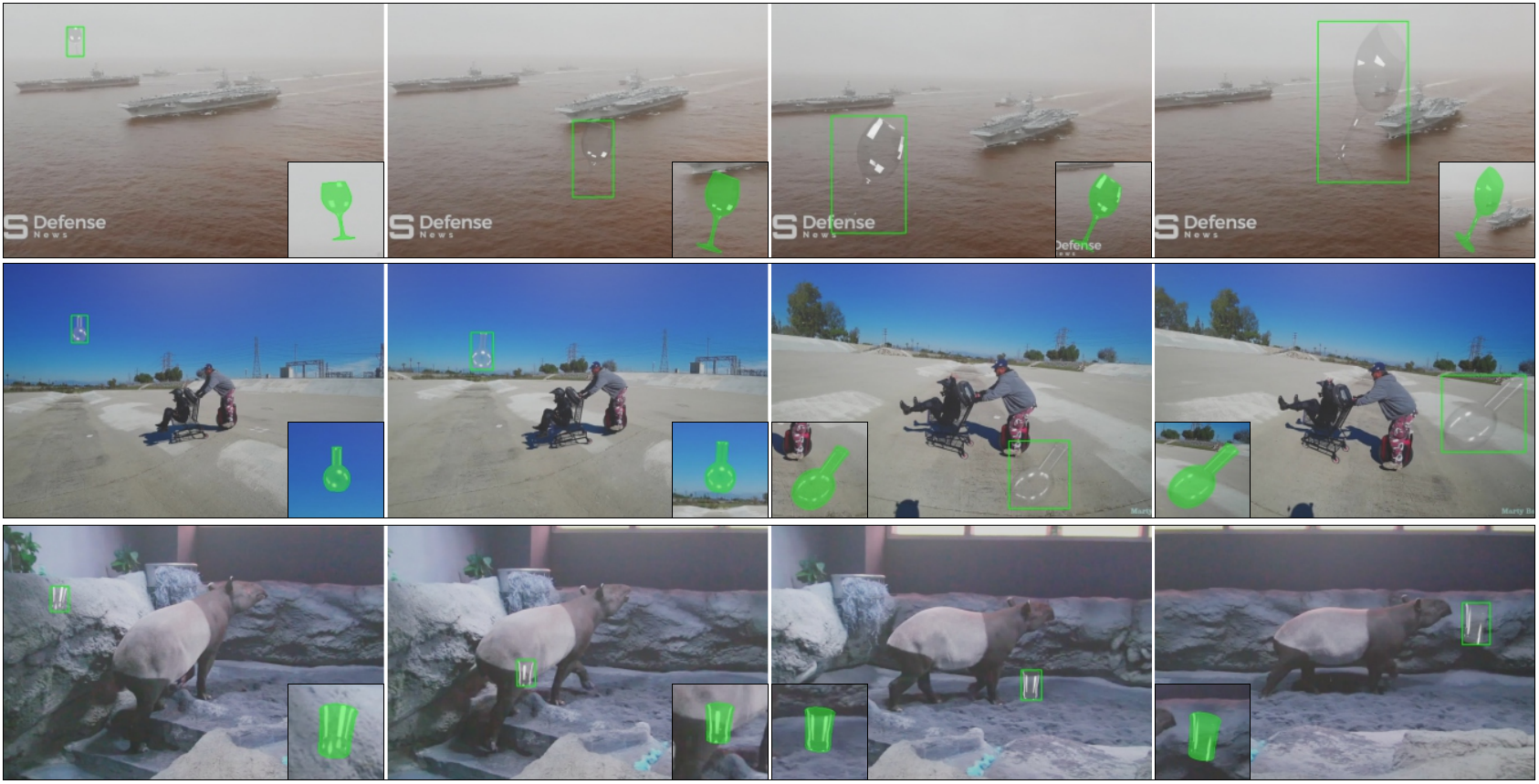}
\caption{Frames from three sequences of the proposed Trans2k dataset. Objects are labeled by bounding boxes and segmentation masks (cropped in square).}
\label{fig:trans2k-qualitative}
\end{figure*}

Figure~\ref{fig:trans2k-qualitative} shows three sequences from the proposed Trans2k dataset. 
Both types of ground-truth annotations are visualized: bounding boxes as well as segmentation masks.
{\color{orange} }
Note that while transparent objects flying over a scene are seldom observed in real life, they convey sufficient level of realism considering the object and immediate surrounding required to train deep models.

\begin{figure*}
\centering
\includegraphics[width=\linewidth]{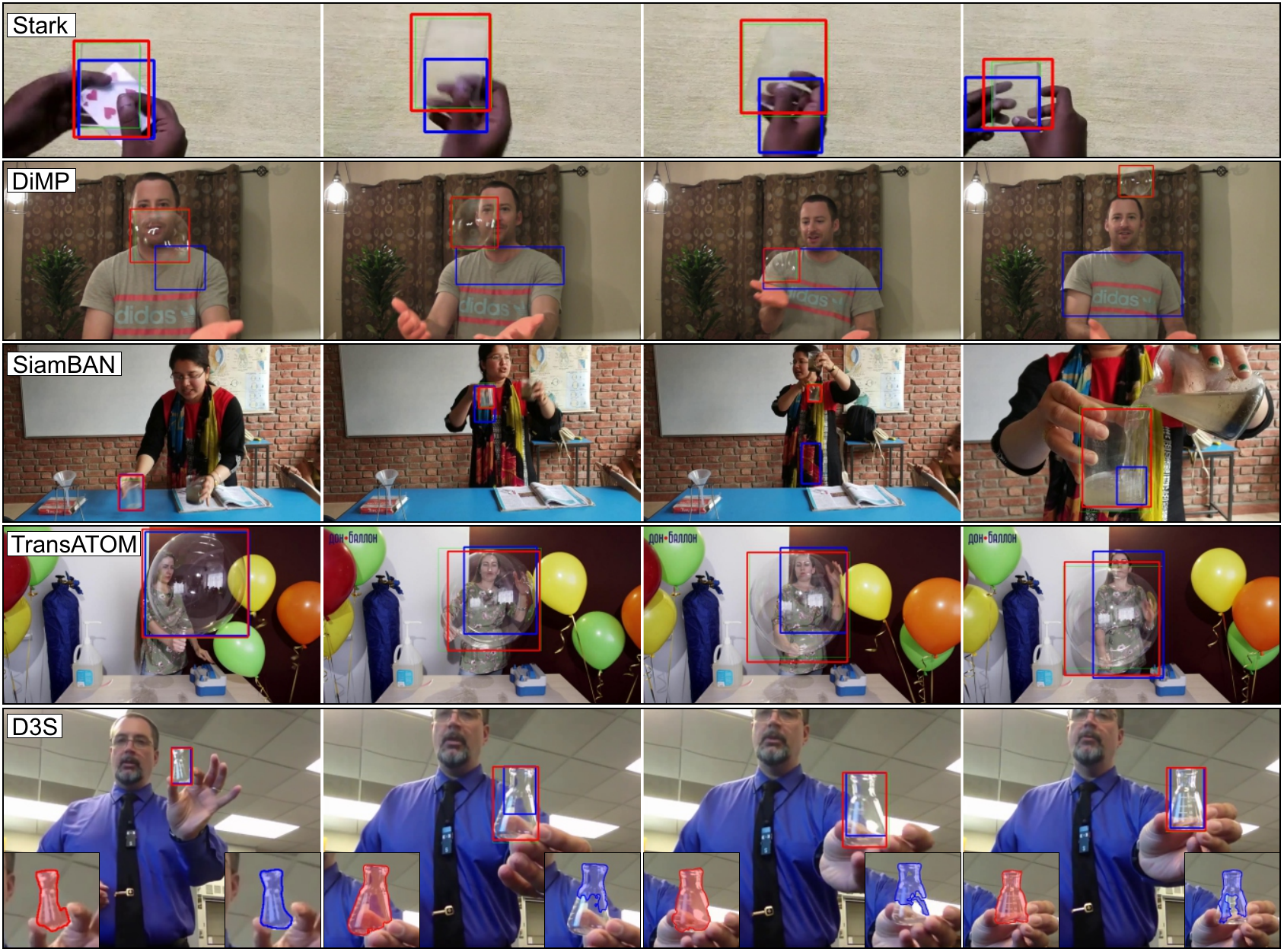}
\caption{Qualitative examples on TOTB dataset. Trackers trained only with opaque object tracking sequences are shown in blue, trackers trained with Trans2k are shown in red and the ground is shown in green.}
\label{fig:tracking-qualitative}
\end{figure*}

In Figure~\ref{fig:tracking-qualitative} we present qualitative comparison of five original trackers and their versions, retrained using Trans2k. 
Trackers trained on Trans2k track the target much more accurately and reliably. 
Since no methodological changes were made to the original trackers, these results clearly demonstrate the effectiveness of the proposed training dataset.

\end{document}